\documentclass[letterpaper]{article} 
\usepackage{aaai25}  
\usepackage{times}  
\usepackage{helvet}  
\usepackage{courier}  
\usepackage[hyphens]{url}  
\usepackage{graphicx} 
\usepackage{natbib}  
\usepackage{caption} 
\usepackage{amsmath}
\usepackage{graphicx}  
\usepackage{booktabs}  
\usepackage{array}     
\usepackage{caption}   
\usepackage{multirow}
\usepackage{amsmath}
\usepackage{amssymb}
\usepackage[table]{xcolor}
\usepackage{url}
\usepackage{xcolor}

\usepackage{algorithm}
\usepackage{algorithmic}

\usepackage{newfloat}
\usepackage{listings}
\DeclareCaptionStyle{ruled}{labelfont=normalfont,labelsep=colon,strut=off} 
\floatstyle{ruled}
\newfloat{listing}{tb}{lst}{}
\floatname{listing}{Listing}
\title{Enhancing Event Reasoning in Large Language Models through \\ Instruction Fine-Tuning with Semantic Causal Graphs}

\author {
    Mazal Bethany\textsuperscript{\rm 1},
    Emet Bethany\textsuperscript{\rm 1},
    Brandon Wherry\textsuperscript{\rm 2},
    Cho-Yu Chiang\textsuperscript{\rm 2},
    Nishant Vishwamitra\textsuperscript{\rm 1},
    Anthony Rios\textsuperscript{\rm 1},
    Peyman Najafirad\textsuperscript{\rm 1}
}
\affiliations {
    \textsuperscript{\rm 1}University of Texas at San Antonio\\
    \textsuperscript{\rm 2}Peraton Labs
}

\usepackage{bibentry}

\begin{document}

\maketitle

\begin{abstract}
Event detection and text reasoning have become critical applications across various domains. While LLMs have recently demonstrated impressive progress in reasoning abilities, they often struggle with event detection, particularly due to the absence of training methods that consider causal relationships between event triggers and types. To address this challenge, we propose a novel approach for instruction fine-tuning LLMs for event detection. Our method introduces Semantic Causal Graphs (SCGs) to capture both causal relationships and contextual information within text. Building off of SCGs, we propose SCG Instructions for fine-tuning LLMs by focusing on event triggers and their relationships to event types, and employ Low-Rank Adaptation (LoRA) to help preserve the general reasoning abilities of LLMs. Our evaluations demonstrate that training LLMs with SCG Instructions outperforms standard instruction fine-tuning by an average of 35.69\% on Event Trigger Classification. Notably, our fine-tuned Mistral 7B model also outperforms GPT-4 on key event detection metrics by an average of 31.01\% on Event Trigger Identification, 37.40\% on Event Trigger Classification, and 16.43\% on Event Classification. We analyze the retention of general capabilities, observing only a minimal average drop of 2.03 points across six benchmarks. This comprehensive study investigates multiple LLMs for the event detection task across various datasets, prompting strategies, and training approaches.
\end{abstract}

%

\section{Introduction}
\label{sec:Introduction}

Event detection, which involves identifying and classifying events within text, has become a crucial application in various domains~\cite{li2022survey}. Its importance is evident in addressing global incidents, tracking public opinions, and analyzing trends across multiple fields~\cite{shin2020cybersecurity}. The roots of event detection can be traced back to the late 1980s, initially focused on identifying terrorism-related events from news sources~\cite{hogenboom2016survey}. Since then, the event detection domain has expanded significantly, highlighting its growing significance.

As users increasingly rely on Large Language Models (LLMs) to support decision-making processes, ensuring the accuracy of these models becomes critical, particularly in high-stakes scenarios~\cite{rawte2023survey}. LLMs are increasingly popular for automatic event detection due to their ability to analyze complex scenarios where understanding context and nuance is crucial~\cite{huang2023textee}. However, leveraging LLMs for automatic event detection faces significant challenges. \textit{First} there is a lack of understanding of how to effectively use LLMs for event detection, requiring studies on their performance in diverse, large-scale scenarios. \textit{Second}, existing LLM training methods do not consider the causal relationship between event triggers (i.e. (the most salient words indicating event types) and event types essential for accurate event detection, necessitating new training strategies.

To address these challenges, we propose a novel approach for instruction fine-tuning LLMs for event detection. Our method introduces Semantic Causal Graphs (SCGs), a new type of directed graph that captures both causal relationships and contextual information within text. Building upon SCGs, we develop Semantic Causal Graph Instructions, a method for generating instruction fine-tuning datasets by extracting causal subgraphs from SCGs, focusing on event triggers and their relationships to event types. We then perform instruction fine-tuning on LLMs using this SCG Instruction event detection dataset, teaching the model to identify causal links behind event classifications by first recognizing event triggers. To preserve the LLM's general language understanding capabilities while adapting it to event detection, we employ the Low-Rank Adaptation (LoRA) technique during fine-tuning.

Our evaluations show that training LLMs with SCG Instructions outperforms standard instruction-fine tuning by an average of 35.69\% on Event Trigger Classification metric. Overall, we find that our fine-tuned Mistral 7B model on SCG Instructions for event detection outperforms GPT-4 on three key event detection evaluation metrics by an average of 31.01\% on Event Trigger Identification, 37.40\% on Event Trigger Classification, and 16.43\% on Event Classification. We additionally analyze our trained LLMs' retention of general reasoning capabilities, observing only a minimal average drop of 2.03 points across six benchmarks compared to the original model's performance. Overall, this study investigates three open-source LLMs and two closed-source LLMs across five event detection datasets, using five prompting strategies and three training strategies. The broader implications of this work suggest that training LLMs on causal relationships may improve task performance, potentially extending beyond event detection to other tasks where understanding causality is crucial.

The main contributions of this work are as follows:
\begin{itemize}
\item We propose Semantic Causal Graphs (SCGs), a novel representation for events that captures both causal relationships and contextual information within text, providing a structured approach to modeling event detection.

\item We develop Semantic Causal Graph Instructions, a method for generating instruction fine-tuning datasets by extracting causal subgraphs from SCGs. This approach focuses on event triggers and their relationships to event types, enabling more effective instruction fine-tuning of LLMs for event detection.

\item We conduct extensive experiments demonstrating that our method outperforms both off-the-shelf LLMs and standard instruction fine-tuning techniques for event detection across multiple datasets and evaluation metrics.
\end{itemize}

\section{Background}
\label{sec:Background}

\subsection{Event Detection}
Events are defined as specific occurrences involving participants~\cite{Doddington2004ACE}. Event detection, a crucial component of information extraction, focuses on identifying event triggers (words or phrases signaling an event) and classifying them into predefined event types~\cite{li2022survey}. This task is fundamental for understanding and extracting structured information from unstructured text. Traditionally, common approaches to event detection include supervised learning with deep learning models~\cite{liao2021learning, wang2019adversarial, tanev2024leveraging} and graph parsing methods~\cite{xie2021event,wan2024dependency,mi2022event}. These approaches have shown promising results in capturing complex event structures and relationships.

Recent research has explored using Large Language Models (LLMs) for event detection, leveraging their powerful language understanding capabilities. This includes techniques such as data augmentation to enrich training datasets~\cite{veyseh2021unleash, chen2024large} and investigating zero-shot/one-shot prompting capabilities~\cite{chen2024large}. However, some work shows that LLMs, even with few-shot examples, can underperform compared to traditional approaches specifically tailored for event detection~\cite{huang2023textee}. This highlights the challenges in adapting general-purpose language models to specialized tasks. Most studies have focused on single datasets and have not extensively explored the potential of retraining LLMs specifically for the nuanced task of event detection across diverse domains and event types.

\subsection{Enhancing LLM Task Performance}
Improving LLM performance on specific tasks faces challenges when pretraining hasn't provided the necessary skills or domain-specific knowledge. Various approaches have been proposed to address this issue, each targeting different aspects of model capabilities. Few-shot learning~\cite{brown2020language} provides the LLM with task examples, leveraging the model's ability to adapt to new tasks with minimal guidance. Retrieval-Augmented Generation (RAG)~\cite{lewis2020retrieval} augments the model with external knowledge, enabling access to information beyond its training data. Chain-of-Thought (CoT) prompting~\cite{wei2022chain} guides LLMs to break down complex problems into intermediate steps, enhancing their reasoning abilities.

More fundamental model training techniques have also been developed to align LLMs with specific tasks or desired behaviors. These include Instruction Tuning~\cite{ouyang2022training}, which fine-tunes models on diverse task instructions, Reinforcement Learning with Human Feedback (RLHF)~\cite{stiennon2020learning}, which optimizes model outputs based on human preferences, and Direct Preference Optimization (DPO)~\cite{rafailov2024direct}, which efficiently adapts model parameters using positive and negative examples. Despite these advancements, there remains a significant gap in explicitly enhancing LLMs' understanding of causal relationships, which could be crucial for tasks requiring such capabilities.

\begin{figure*}[h]
\centering
\includegraphics[width=0.9\textwidth]{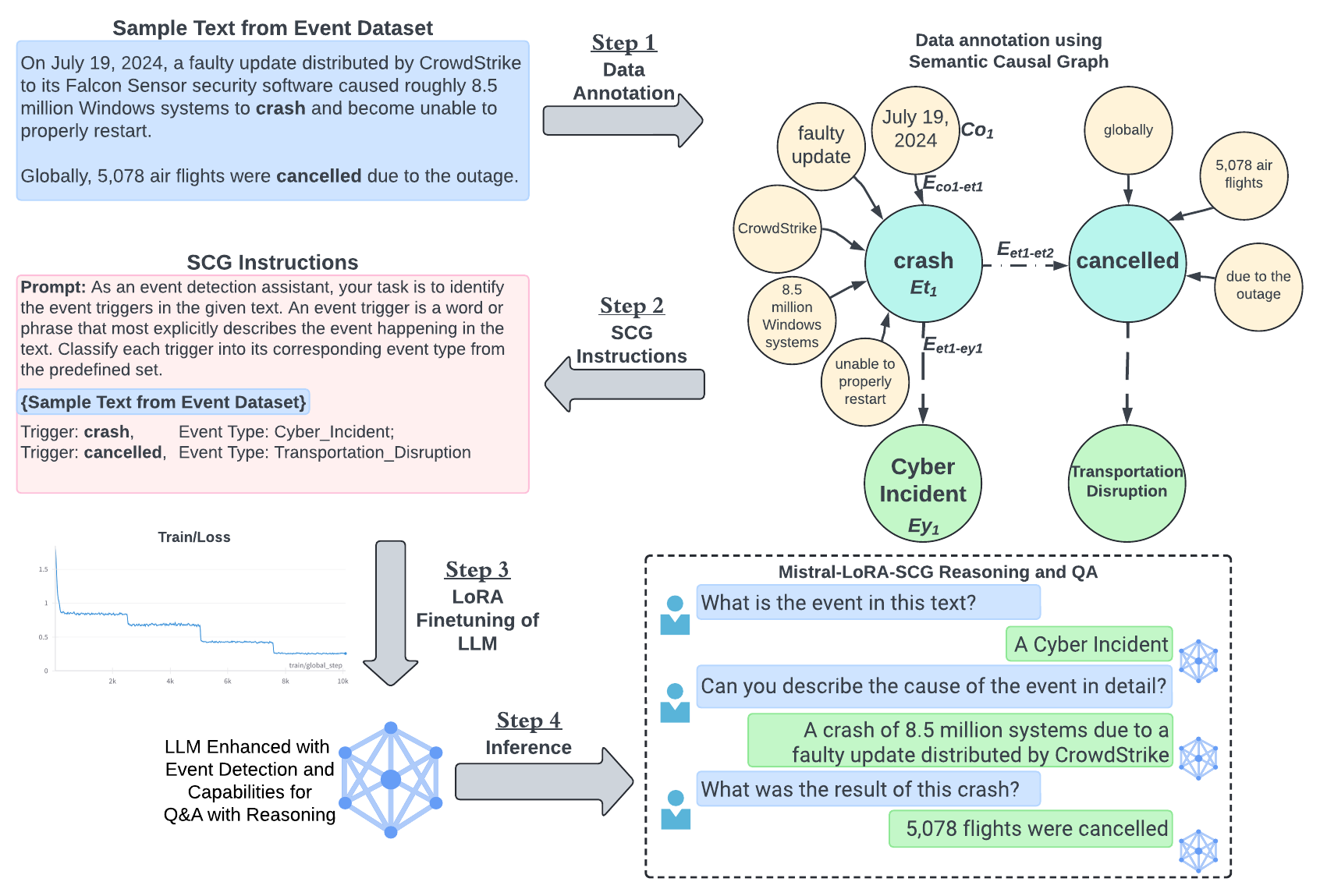}
\caption{SCG Instructions for instruction fine-tuning. By including causality in the instruction fine-tuning dataset, the LLM learns the causal relationship between causality nodes and classification types.}
\label{fig:SCG Instructions_vs_standard}
\end{figure*}

\section{Method}
\label{sec:Method}

We propose a novel approach for instruction fine-tuning LLMs to enhance their performance in event detection. We first introduce Semantic Causal Graphs (SCGs), a new type of directed graph that captures both the contextual (i.e., temporal, spatial, situational, etc.) information within the text, as well as the event trigger content that affects event classification. Building upon SCGs, we develop a Semantic Causal Graph Instruction dataset to fine-tune LLMs, aiming to enhance the model's performance on event detection. SCG Instructions are created by extracting the causal subgraph from the SCG, focusing on event trigger nodes and their relationships, where these nodes serve as key causal elements that lead to event classifications. This approach during the fine-tuning process teaches the LLM to accurately identify event triggers as critical causal elements, thereby enhancing the LLM's event detection abilities. This is achieved while preserving its general language capabilities through the use of the Low-Rank Adaptation (LoRA) technique. Figure \ref{fig:SCG Instructions_vs_standard} illustrates this process, showing the steps from creating SCGs to extracting causal subgraphs for SCG Instructions, and finally fine-tuning the LLM using these instructions.

\subsection{Semantic Causal Graph Definition}

A Semantic Causal Graph (SCG) is a directed graph $G = (V, E)$ representing a causal structure for an input $Text$ and a corresponding event label $L$. The set of nodes $V$ is partitioned into three disjoint subsets: Context nodes $Co = \{co_1, \ldots, co_m\}$, Event trigger nodes $Et = \{et_1, \ldots, et_n\}$, and Event type nodes $Ey = \{ey_1, \ldots, ey_p\}$.

\noindent To better understand the components of this graph structure and their roles in representing causal relationships within the text for event detection, we expand on each type of node in detail:

\noindent \textbf{Context nodes ($Co$):} These represent background information, settings, or conditions under which events occur. They provide essential details for understanding the broader scenario, including temporal, spatial, and situational information. Context nodes can connect to event trigger nodes, providing the backdrop against which causal chains of events unfold.

\noindent \textbf{Event Trigger nodes ($Et$):} These represent specific actions, occurrences, or factors that drive the narrative or scenario forward, indicating the presence of an event. They are typically verbs, verb phrases, or key elements denoting actions, changes, or influential factors. Event trigger nodes can connect to other event trigger nodes, showing how one event can lead to another. Each event trigger node must have exactly one outgoing edge to an event type node, representing the direct causal relationship between the event trigger and the event type.

\noindent \textbf{Event Type nodes ($Ey$):} These represent the event classifications. Event-type nodes are the endpoints of causal chains within the graph, capturing the event types resulting from the event triggers.

Context nodes ($Co$) and event trigger nodes ($Et$) are derived from the input $Text$. Event type nodes ($Ey$) are derived from the event label $L$.

\noindent The set of directed edges $E$ consists of three types of connections:

\begin{itemize}
\item $E_{co\text{-}et} \subseteq (Co \times Et)$: Context-to-trigger relationships. This relationship represents how context sets the stage for event triggers that lead to the occurrence of events.
\item $E_{et\text{-}et} \subseteq (Et \times Et)$: Trigger-to-trigger relationships. This relationship captures the causal chain between events, showing how one event trigger can lead to or influence another, thus modeling the sequential or interconnected nature of events in the text.
\item $E_{et\text{-}ey} \subseteq (Et \times Ey)$: Trigger-to-type relationships. This relationship links event triggers to their corresponding event types, representing the classification or categorization of events based on their triggering actions or occurrences.
\end{itemize}

Each trigger node requires precisely one outgoing edge connecting to an event type node $ey_j$. Graph $G$ serves as a structural representation of contextual data, event triggers, and their relationships to event types. This structure is derived from the input $Text$ and its associated event label $L$, capturing the causal and semantic connections within the given information. To build this graph, annotators create nodes representing the context, event triggers, and event types, as well as define the edges that capture the relationships between these elements within the text. This process involves decomposing the text into triggers and context and then labeling the event type. This comprehensive annotation encapsulates information necessary for building the SCG for event detection.

\subsection{SCG Instruction Dataset}

The SCG allows us to model complex interactions within text, focusing on the causal chains that lead to specific outcomes. We focus on a simplified version of the SCG, which we call the causal subgraph, that emphasizes the causal relationships most relevant to identifying event types. In this focused representation, Event Trigger nodes serve as the primary nodes, and their causal relationships to Event Type nodes form the edges. This approach is crucial because it guides the model to learn the most relevant causal aspects for event detection, rather than potentially being distracted by peripheral contextual information. Non-essential information is known to cause models to learn spurious relationships, which can lead to reduced model generalizability~\cite{ye2024spurious}.

We then transform the causal subgraph of the SCG into an instruction-tuning dataset. This allows the LLM to learn the direct causal relationships between triggers and event types. We transform this structured data into an instruction-tuning format that explicitly includes the causal trigger before the event type, modeling the probability of an event trigger given the input text $P(Et|Text)$ and the probability of an event type given a trigger $P(Ey|Et)$, ultimately allowing it to compute the overall conditional probability $P(Ey|Text)$. We can then express the causal relationships in the SCG using the following probabilistic formulation:

\begin{equation}
\label{eq:conditional}
P(Ey|Text) = \sum_{Et} P(Ey|Et)P(Et|Text)
\end{equation}

The key distinction between SCG Instructions and standard instruction tuning lies in the sequential presentation of causal information. In standard instruction tuning for event detection, the model is typically trained to directly predict the event type $Ey$ given the input $Text$, directly learning $P(Ey|Text)$. In contrast, SCG Instructions has the model first identify the event trigger $Et$ and then use this information to classify the event type $Ey$, thus decomposing the problem into learning $P(Ey|Et)$ and $P(Et|Text)$. This approach aligns with the causal formulation $P(Ey|Text) = \sum_{Et} P(Ey|Et)P(Et|Text)$ introduced earlier, encouraging the model to explicitly consider the intermediate causal step of identifying the event trigger before making the final classification. 

\begin{table*}[]
\centering
\resizebox{.99\textwidth}{!}{
\begin{tabular}{@{}ll|ccc|ccc|ccc|ccc|ccc|ccc@{}}
\toprule
\multirow{4}{*}{}                & \multirow{4}{*}{\textbf{Model}}    & \multicolumn{3}{c|}{\multirow{3}{*}{\textbf{MAVEN}}} & \multicolumn{3}{c|}{\multirow{3}{*}{\textbf{FewEvent}}} & \multicolumn{3}{c|}{\multirow{3}{*}{\textbf{M$^2$E$^2$}}} & \multicolumn{3}{c|}{\multirow{3}{*}{\textbf{CASIE}}} & \multicolumn{3}{c|}{\multirow{3}{*}{\textbf{MLEE}}} & \multicolumn{3}{c}{\multirow{3}{*}{\textbf{Average}}} \\
                                 &                           & \multicolumn{3}{l|}{}                       & \multicolumn{3}{l|}{}                          & \multicolumn{3}{l|}{}                      & \multicolumn{3}{l|}{}                       & \multicolumn{3}{l|}{}                       & \multicolumn{3}{l}{}                         \\
                                 &                           & \multicolumn{3}{c|}{(General)}              & \multicolumn{3}{c|}{(General)}                & \multicolumn{3}{c|}{(News)}                & \multicolumn{3}{c|}{(Cybersecurity)}       & \multicolumn{3}{c|}{(Biomedical)}           & \multicolumn{3}{c}{}                \\
                                 &                           & EC           & TI           & TC           & EC            & TI            & TC            & EC           & TI           & TC          & EC           & TI           & TC           & EC            & TI            & TC            & EC            & TI            & TC           \\ 
\midrule

\multirow{3}{*}{Prompt learning}      & GPT-4 (zero-shot)         & 47.14        & 49.89        & 19.13        & 50.58         & 35.87         & 23.39         & 53.33        & 13.33        & 10.29       & 32.60        & 7.44         & 6.89         & 19.13        & 22.51         & 19.04         & 40.56         & 25.81         & 15.75        \\
                                 & GPT-4 (six-shot)          & 58.84        & 56.96        & 31.01        & 52.80         & 35.60         & 21.47         & 71.94        & 59.29        & 59.29       & 61.65        & 15.78        & 15.08        & 66.77        & 43.51         & 38.62         & 62.40         & 42.23         & 33.09        \\
                                 & GPT-4 (six-shot RAG)      & \textbf{68.90} & \textbf{58.85} & \textbf{42.80} & \textbf{65.20} & \textbf{45.20} & \textbf{40.80} & \textbf{83.56} & \textbf{63.37} & \textbf{63.37} & \textbf{92.19} & \textbf{32.04} & \textbf{31.50} & \textbf{77.01} & \textbf{50.00} & \textbf{43.60} & \textbf{77.37} & \textbf{49.89} & \textbf{44.41}
        \\ \midrule

\multirow{3}{*}{Prompt learning} & Mistral (zero-shot)       & 29.17        & 20.50        & 2.31         & 18.28         & 12.21         & 2.63          & 45.37        & 2.43         & 1.42        & 39.92        & 3.30         & 1.02         & 45.77        & 7.90          & 0.72          & 35.70         & 9.27          & 1.62         \\
                                 & Mistral (six-shot)        & 24.05        & 29.45        & 8.41         & 23.48         & 6.33          & 2.95          & 74.84        & 46.89        & 45.93       & 49.00        & 5.98         & 4.93         & 51.29        & 27.02         & 18.29         & 44.53         & 23.13         & 16.10        \\
                                 & Mistral (six-shot RAG)    & 43.97        & 38.95        & 19.82        & 48.66         & 30.38         & 25.38         & 69.10        & 46.90        & 46.90       & 68.17        & 12.13        & 10.85        & 62.72        & 26.11         & 20.76         & 58.52         & 30.89         & 24.74        \\ \cmidrule{1-1} 

\multirow{2}{*}{Fine-tuning}     & Mistral-LoRA-Instruct     & 33.84        & 15.94        & 13.28        & 83.99         & 46.00         & 43.60         & 89.30        & 85.82        & 85.06       & 50.02        & 25.78         & 22.84         & 56.94        & 43.49         & 37.55         & 62.82         & 43.41         & 40.47        \\
                                 & Mistral-LoRA-SCG Instruct & \textbf{92.65} & \textbf{73.59} & \textbf{61.28} & \textbf{94.74} & \textbf{63.61} & \textbf{61.47} & \textbf{94.54} & \textbf{85.70} & \textbf{85.59} & \textbf{96.70} & \textbf{47.49} & \textbf{46.81} & \textbf{71.75} & \textbf{56.56} & \textbf{49.96} & \textbf{90.08} & \textbf{65.39} & \textbf{61.02}
        \\ \midrule

\multirow{3}{*}{Prompt learning} & Llama 3 (zero-shot)       & 32.16        & 34.33        & 8.58         & 34.00         & 24.80         & 12.40         & 38.49        & 5.95         & 3.97        & 43.13        & 2.75         & 2.22         & 6.54         & 5.15          & 0.69          & 30.86         & 14.60         & 5.57         \\
                                 & Llama 3 (six-shot)        & 38.16        & 46.74        & 13.44        & 45.60         & 24.80         & 14.80         & 61.78        & 43.17        & 42.77       & 57.19        & 9.07         & 7.41         & 67.88        & 36.50         & 23.98         & 54.12         & 32.06         & 20.48        \\
                                 & Llama 3 (six-shot RAG)    & 64.73        & 54.01        & 31.90        & 62.40         & 40.00         & 34.80         & 55.34        & 50.59        & 43.48       & 93.70        & 21.70         & 20.69         & 76.20        & 36.84         & 28.18         & 70.47         & 40.63         & 31.81        \\ \cmidrule{1-1} 

\multirow{2}{*}{Fine-tuning}     & Llama 3-LoRA-Instruct     & 72.65        & 53.89        & 42.35        & 77.86         & 53.81         & 45.64         & 94.75        & 79.23        & 78.89       & 31.60        & 15.37         & 15.10         & 84.94        & 52.69         & 47.08         & 72.36         & 51.00         & 45.81        \\
                                 & Llama 3-LoRA-SCG Instruct & \textbf{89.76} & \textbf{73.11} & \textbf{59.55} & \textbf{91.04} & \textbf{60.42} & \textbf{55.72} & \textbf{95.93} & \textbf{83.60} & \textbf{83.60} & \textbf{57.56} & \textbf{42.63} & \textbf{40.04} & \textbf{85.76} & \textbf{58.96} & \textbf{51.71} & \textbf{84.01} & \textbf{63.74} & \textbf{58.12}
        \\ \midrule

\multirow{3}{*}{Prompt learning} & Gemma (zero-shot)         & 7.60        & 7.31        & 0.68        & 20.88         & 3.26         & 1.86         & 22.37        & 7.34        & 6.42       & 21.94        & 1.12         & 0.86         & 4.02        & 0.35         & 0.00         & 15.36         & 3.88         & 1.96        \\
                                 & Gemma (six-shot)          & 12.25        & 11.63        & 2.91        & 13.10         & 8.28         & 1.38         & 13.87        & 11.68        & 10.95       & 25.91        & 1.14         & 1.00         & 36.16        & 18.44         & 9.93         & 20.26         & 10.23         & 5.23        \\
                                 & Gemma (six-shot RAG)      & 17.53        & 11.71        & 5.86        & 3.76         & 1.50         & 0.75         & 5.34        & 3.15        & 3.15       & 24.54        & 3.39         & 2.80         & 31.17        & 8.38         & 7.37         & 16.47         & 5.63         & 3.99        \\ \cmidrule{1-1} 

\multirow{2}{*}{Fine-tuning}     & Gemma-LoRA-Instruct       & 59.42        & 37.16        & 29.57        & 58.43         & 47.28         & 45.61         & \textbf{93.16}        & 75.97        & 75.96       & 61.81        & 27.25         & 26.69         & 66.83        & 40.70         & 38.24         & 67.93         & 45.67         & 43.21        \\
                                 & Gemma-LoRA-SCG Instruct   & \textbf{94.61}        & \textbf{70.85}        & \textbf{59.21}        & \textbf{94.11}         & \textbf{58.87}         & \textbf{56.01}         & 84.29        & \textbf{81.78}        & \textbf{81.77}       & \textbf{75.14}        & \textbf{36.86}         & \textbf{33.59}         & \textbf{87.89}        & \textbf{53.95}         & \textbf{48.95}         & \textbf{87.21}         & \textbf{60.46}         & \textbf{55.91}        \\
\bottomrule
\end{tabular}}
\caption{Performance comparison across different models and datasets. F1 scores are reported. Bold numbers indicate the best performance within each model architecture. Results show that LLMs trained on SCG Instructions outperform other LLM strategies.}
\label{tab:table_1_experiments}
\end{table*}

\subsection{Fine-Tuning with LoRA}
To efficiently adapt LLMs for event detection while preserving pre-trained knowledge, we employ Low-Rank Adaptation (LoRA)~\cite{hu2022lora}. LoRA applies a low-rank decomposition to the weight matrices in the transformer layers, significantly reducing the number of trainable parameters.
Given a pre-trained weight matrix $W_0 \in \mathbb{R}^{d \times k}$, LoRA represents its adaptation as:
\begin{equation}
W = W_0 + \Delta W = W_0 + BA
\end{equation}
where $B \in \mathbb{R}^{d \times r}$ and $A \in \mathbb{R}^{r \times k}$ are learnable matrices, and $r \ll min(d,k)$ is the rank of the decomposition. The forward pass through the adapted layer becomes:
\begin{equation}
h = W_0 x + \Delta W x = W_0 x + BA x
\end{equation}
This approach reduces the number of trainable parameters from $d \times k$ to $r \times (d + k)$, leading to lower memory requirements and faster training times. By freezing the pre-trained weights $W_0$ and only updating the low-rank matrices $A$ and $B$, LoRA acts as a strong regularizer, preventing overfitting to the limited event detection data.

\section{Data}
\label{sec:Data}

To evaluate LLMs on event detection, we utilize five diverse open-source datasets processed by the TextEE benchmark method~\cite{huang2023textee}. These datasets were carefully selected to represent a range of domains, input complexities, and output complexities, allowing for a comprehensive assessment of LLM performance across various scenarios. The datasets, listed in order from general to specific domains, are:

\begin{itemize}
\item MAVEN~\cite{wang2020maven}: General domain from Wikipedia (28734 training, 5925 test samples). Features 168 event types with single-sentence, multi-trigger input structure, representing high output complexity.

\item FewEvent~\cite{deng2020meta}: General domain from Wikipedia and Freebase (7579 training, 2541 test samples). Contains 100 event types with single-sentence, single-trigger input structure, also exhibiting high output complexity.

\item CASIE~\cite{satyapanich2020casie}: Cybersecurity news domain (1047 training, 218 test samples). Contains 5 event types with multi-sentence, multi-trigger input structure, representing low output complexity but high input complexity.

\item M$^2$E$^2$~\cite{li-etal-2020-cross}: News domain from Voice of America news (4211 training, 901 test samples). Includes 8 event types with single-sentence, multi-trigger input structure, offering low output complexity. Notably, this dataset has a significant class imbalance, with more than 80\% of the sentences not containing an event trigger.

\item MLEE~\cite{pyysalo2012event}: Biomedical domain focusing on angiogenesis (199 training, 42 test samples). Contains 29 event types with multi-sentence, multi-trigger input structure, presenting medium output complexity and high input complexity.

\end{itemize}





We chose these datasets to rigorously evaluate LLM performance across a spectrum of event detection scenarios. Our selection spans multiple domains, including general, cybersecurity, news, and biomedical, allowing us to assess LLM capabilities in varied subject areas with specialized vocabularies. The datasets' complexity stems from two key factors: diverse input structures and varying output spaces. Input complexity ranges from simple single-sentence, single-trigger cases to more intricate multi-sentence scenarios with multiple event triggers. Output complexity varies significantly, with the number of possible event types ranging from as few as 5 to as many as 168.

\section{Experimental Evaluation}
\label{sec:Experiments}

We begin our evaluation of our proposed SCG Instructions method for fine-tuning LLMs for event detection by first comparing against other LLM methods of performing event detection. This initial comparison aims to demonstrate the significant performance improvements achieved by our approach in addressing the challenges faced by LLMs on event detection. We then evaluate the general language capabilities of some of the models trained on SCG Instructions to show that these models retain their general language capabilities. This assessment is crucial as it verifies the versatility of our fine-tuned models, ensuring their utility extends beyond event detection to other text analysis tasks. Finally, we compare our event detection LLMs against traditional non-LLM approaches for performing event detection. This final comparison serves to illustrate how our method narrows the performance gap between general-purpose language models and these task-specific models.

\begin{table*}[]
\centering
\resizebox{.85\textwidth}{!}{
\renewcommand{\arraystretch}{1.1}%
\begin{tabular}{@{}lrrrrrrr@{}}
\toprule
Model                                  & \textbf{ARC}   & \textbf{HellaSwag} & \textbf{TruthfulQA} & \textbf{MMLU}  & \textbf{Winogrande} & \textbf{GSM8K} & \textbf{Average} \\ \midrule
Llama 3            & 62.12 & 78.80     & 51.66      & 65.63 & 75.61      & 76.04 & 68.31   \\ 
Llama   3-LoRA-SCG Instruct (MAVEN)    & 56.14 & 77.68     & 49.61      & 62.58 & 68.98      & 62.77 & 62.96   \\
Llama   3-LoRA-SCG Instruct (FewEvent) & 61.18 & 79.14     & 48.57      & 64.78 & 71.51      & 65.50 & 65.11   \\
Llama   3-LoRA-SCG Instruct (M2E2)     & 63.31 & 80.44     & 52.38      & 64.74 & 73.01      & 71.72 & 67.60   \\
Llama   3-LoRA-SCG Instruct (CASIE)    & 64.42 & 82.61     & 51.10      & 64.54 & 74.82      & 67.63 & 67.52   \\
Llama   3-LoRA-SCG Instruct (MLEE)     & 63.40 & 81.71     & 51.56      & 65.20 & 76.48      & 70.96 & 68.22   \\ 
\bottomrule
\end{tabular}}
\caption{General LLM capabilities measured on several popular benchmarks. We compare the original Llama 3 Instruct model against versions of the Llama 3-LoRA-SCG Instruct that we trained for event detection.}
\label{tab:general_LLM}
\end{table*}

\subsection{Evaluation Metrics}

The primary metrics used to evaluate the performance of event detection methodologies are event classification (EC), event trigger identification (TI), and event trigger classification (TC). EC directly measures the system's ability to correctly classify the events occurring in a given text, which is the primary goal of event detection. TI focuses on identifying the specific words in the text that cause the event. TC combines both EC and TI, requiring the system to classify the event and identify the corresponding trigger word correctly. As a result, TC will always be lower than or equal to both EC and TI.

\subsection{LLMs for Event Detection}
\label{sec:instruct_tuning_eval}

The primary objective of our initial experiments is to evaluate the effectiveness of fine-tuning with SCG Instructions compared to other LLM-based approaches for Event Detection. We conduct a comparative analysis involving GPT-4 Turbo~\cite{openai_gpt4turbo0409_2024}, Mistral 7B~\cite{jiang2023mistral}, Llama 3 8B \cite{touvron2023llama}, and Gemma 2B \cite{team2024gemma} across three scenarios: zero-shot, six-shot, and six-shot augmented with RAG. In the zero-shot setting, we provide only a description of the event detection task and the types of events to be classified. The six-shot scenario builds upon this by including six randomly selected input-output examples. Finally, the six-shot RAG approach refines the selection process further by choosing input-output examples based on the cosine similarity of the test text embedding to the input text embedding rather than random selection. We additionally compare training on our SCG Instruction fine-tuning dataset methodology to training on standard instruction fine-tuning datasets for event detection to show how incorporating the causal graph in fine-tuning improves the performance of the LLM in Event Detection. For the fine-tuning methods, we use the LoRA implementation from the Unsloth library~\cite{unsloth} to implement LoRA for doing instruction fine-tuning on LLMs, using cross-entropy loss for training the LoRA parameters. We use open source implementations of the Gemma 2B~\cite{team2024gemma}, Mistral 7B~\cite{jiang2023mistral}, and Llama 3 8B~\cite{llama3} architectures. For the Gemma 2B model, we use the pretrained gemma-2b~\cite{google_gemma2b_2024} on Hugging Face. For the Mistral 7B model, we utilize the base Mistral 7B model, Mistral-7B-v0.2-hf~\cite{alpindale_mistral7bv02hf_2024} on Hugging Face. For the Llama 3 8B model, we train on top of the Meta-Llama-3-8B-Instruct model~\cite{Meta-Llama-3-8B-Instruct} from Hugging Face. We denote the LLMs trained with the standard instruct with ``-LoRA-Instruct'' and the models trained with SCG instructions with ``-LoRA-SCG Instruct''. Complete details about the experimental setup are provided in the Appendix.

We present the results of these experiments in Table~\ref{tab:table_1_experiments}. We find on the prompt learning methods, six-shot RAG outperformed the zero-shot and six-shot scenarios across all models, which is to be expected. However, we also see that fine-tuning provides better performance over the the prompt learning methods. As seen by the results in Table~\ref{tab:table_1_experiments}, Event detection is a challenging task for LLMs, where we observe a TC score from GPT-4 of just 44.41 when prompting it with six-shot RAG. We see that incorporating examples with six-shot prompting improves the results, and using RAG to find more relevant samples to the input text improves the results even further across most models and datasets. Comparing the ``-LoRA-Instruct'' to the ``-LoRA-SCG Instruct'' models, we see a large increase in performance across the EC, TI, and TC metrics across all models. This shows that incorporating the causal subgraph in the training process greatly enhances the LLM's understanding and performance in the event detection task.

To further demonstrate that using LoRA is unlikely to negatively impact the performance of event detection when training with our SCG Instructions method, we additionally trained Mistral with full parameter fine-tuning on the M$^2$E$^2$ dataset. We found that fine-tuning with LoRA yielded slightly better performance than full fine-tuning. The LoRA model achieved an EC of 95.99, TI of 85.27, and TC of 85.16, while the full fine-tuning model had an EC of 83.63, TI of 82.49, and TC of 82.49.


\begin{table*}[]
\centering
\resizebox{.90\textwidth}{!}{
\begin{tabular}{@{}l|c|cc|cc|cc|cc|cc|cc@{}}
\toprule
\multirow{2}{*}{Model}    & \multirow{2}{*}{LLM} & \multicolumn{2}{c|}{\textbf{MLEE}} & \multicolumn{2}{c|}{\textbf{FewEvent}} & \multicolumn{2}{c|}{\textbf{M$^2$E$^2$}} & \multicolumn{2}{c|}{\textbf{CASIE}} & \multicolumn{2}{c|}{\textbf{MAVEN}} & \multicolumn{2}{c}{\textbf{Average}} \\
                          &                                 & TI           & TC                & TI           & TC             & TI           & TC             & TI           & TC                & TI           & TC                & TI              & TC               \\ 
\midrule
TagPrime-C                & No                              & 82.60        & 78.20             & 67.20        & 65.60           & 53.10        & 51.00          & 44.90        & 44.70             & 74.70        & 66.10             & 64.50           & 61.12            \\
CEDAR                     & No                              & 71.00        & 65.50             & 66.90        & 52.10           & 50.90        & 48.00          & 68.70        & 67.60             & 76.50        & 54.50             & 66.80           & 57.54            \\
DEGREE                    & No                              & 74.00        & 70.40             & 67.90        & 65.50           & 50.40        & 48.30          & 61.50        & 61.30             & 76.20        & 65.50             & 66.00           & 62.20            \\ \midrule
Mistral-LoRA-SCG Instruct & Yes                             & 56.56        & 49.96             & 63.61        & 61.47           & 85.70        & 85.59          & 47.49        & 46.81             & 73.59        & 61.28             & 65.39           & 61.02            \\
Llama 3-LoRA-SCG Instruct & Yes                             & 58.96        & 51.71             & 60.42        & 55.72           & 83.60        & 83.60          & 42.63        & 40.04             & 73.11        & 59.55             & 63.74           & 58.12            \\
Gemma-LoRA-SCG Instruct   & Yes                             & 53.95        & 48.95             & 58.87        & 56.01           & 81.78        & 81.77          & 36.86        & 33.59             & 70.85        & 59.21             & 60.46           & 55.91            \\
\bottomrule
\end{tabular}}
\caption{Comparison of top performing event detection methods against LLMs fine tuned on SCG Instructions. Our SCG Instructions help to narrow the gap between general-purpose LLMs and task-specific models.}
\label{tab:comparison_models}
\end{table*}

\subsection{SCG Fine-tuned LLM General Language Capabilities}

To assess the general reasoning capabilities of LLMs fine-tuned on our custom event detection dataset, we evaluated their performance on several key LLM benchmarks: ARC~\cite{clark2018think}, HellaSwag~\cite{zellers2019hellaswag}, MMLU~\cite{hendrycks2021measuring}, TruthfulQA~\cite{lin-etal-2022-truthfulqa}, Winogrande~\cite{sakaguchi2021winogrande}, and GSM8k~\cite{cobbe2021gsm8k}. These standardized datasets evaluate various aspects of language understanding, reasoning, and factual accuracy across different domains and tasks. We used the ElutherAI implementation~\cite{lm_evaluation_harness} for these experiments. Table \ref{tab:general_LLM} presents the results for the different Llama 3 8B models that were fine-tuned with SCG Instructions. The original Meta-Llama-3-8B-Instruct model~\cite{Meta-Llama-3-8B-Instruct} served as our baseline to compare to.

The original Meta-Llama-3-8B-Instruct model generally outperformed most other models we tested across the benchmarks. In particular, the model trained on SCG Instructions with Genia2011 data slightly outperformed the original model on average, though the difference was minimal.
We observed that models trained on SCG Instructions datasets with fewer samples (MLEE, CASIE) performed better than those with more samples (MAVEN). We attribute this decreased performance with larger training sets to the catastrophic forgetting problem often encountered in LLM training for specific tasks \cite{zhai2024investigating, liu2024more, lin2023speciality}.
When fine-tuning an LLM on only event detection-related data, it's possible for the model to start losing some of its general language capabilities due to this focused training. MAVEN is the largest of the datasets we tested, and we observed a more significant performance drop compared to models trained on other event detection datasets.
However, despite this drop in performance, the general language capabilities remain strong. For context, the base Llama-2 70B Instruct model has an average score of 62.4 across the same benchmarks, compared to our lowest scoring model (Llama 3-LoRA-SCG Instruct (MAVEN)), which had an average score of 62.96, indicating that our trained models retain the general reasoning abilities to appropriately respond to other types of queries.

\subsection{SCG Instruction Fine-tuned LLMs vs. Deep Learning Models}

We next compare our LLMs that were fine-tuned with SCG Instructions against three of the top performing non-LLM based event detection models DEGREE~\cite{hsu2022degree}, CEDAR~\cite{li2023glen}, and TagPrime-C~\cite{hsu2023tagprime}. This comparison is motivated by the challenges LLMs face in event detection tasks, as outlined in our background section and in Table~\ref{tab:table_1_experiments}. The results of this experiment are presented in  Table \ref{tab:comparison_models}. While task-specific models may still outperform overall, our results demonstrate that we are closing the performance gap. These findings show the potential of general-purpose LLMs in specialized domains and highlight the substantial advancements being made in adapting these versatile models to targeted tasks, bringing their performance increasingly in line with specialized approaches. The details of the implementation of these models are provided in the Appendix.

Interestingly, our LLMs showed significant performance variations across these datasets. Notably, it underperformed significantly on the MLEE and CASIE datasets while outperforming on the M2E2 dataset compared to baseline non-LLM approaches. These disparities may be attributed to several factors. The underperformance on MLEE potential stems from its highly specialized biomedical domain and extremely small dataset size (199 training samples), which challenges the LLM's broad but potentially shallow knowledge in specific fields. Similarly, CASIE's cybersecurity focus and relatively small dataset (1047 training samples) may have contributed to the LLM's struggles. In contrast, the M2E2 dataset, with its general news domain and larger size (4211 training samples), aligns well with the LLM's strengths in broad knowledge and contextual understanding. While LLMs show promise in certain scenarios, they still face challenges in highly specialized or data-scarce domains where task-specific methods with carefully engineered features may hold an advantage. 

\begin{figure}[b]
\centering
\includegraphics[width=1\columnwidth]{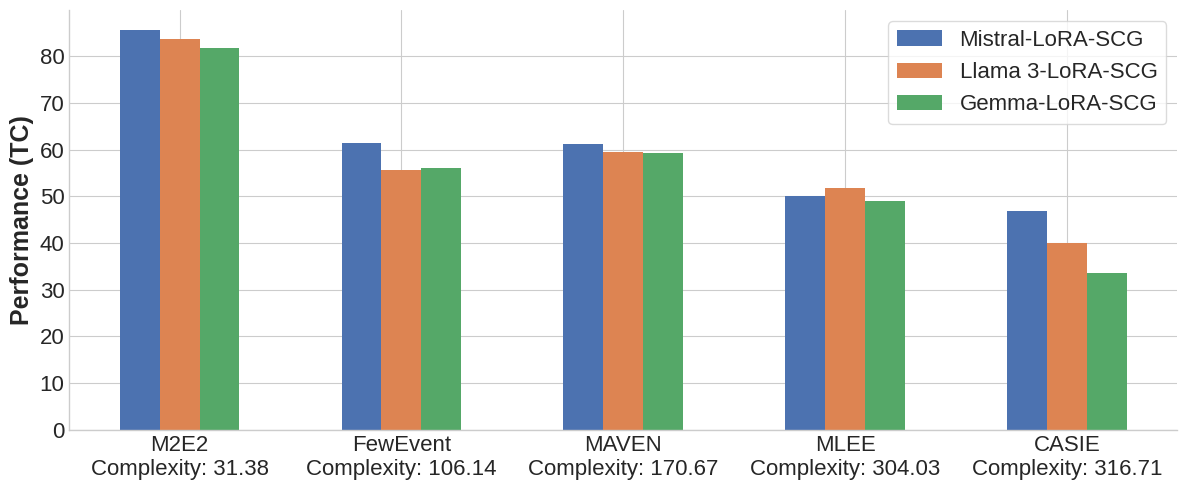}
\caption{Model performance (TC) across datasets, sorted by composite complexity score.}
\label{fig:dataset_complexity}
\end{figure}

\noindent \textbf{Ablation Study.} To demonstrate that our model effectively learned the causal relationship between event triggers and event types, rather than relying on peripheral contextual information, we conducted an additional study. We extracted and replaced context words, including temporal and spatial information, in the test set data across all five datasets, while keeping event trigger words unchanged. This process resulted in modified texts that maintained the same events but with altered contextual information. We applied this experimental procedure to each dataset and found that the average TC score dropped from the original test set to the new test set by 3.58 points. These results showed minimal deviation from the performance on the original test set data. This consistency in performance between the original and modified test sets suggests that our model, including SCG Instruct, has indeed learned to focus on the causal relationship between event triggers and event types, rather than being overly reliant on contextual cues. These findings support the effectiveness of our approach in enhancing the model's ability to identify and utilize causally relevant information for event detection tasks, while demonstrating robustness to changes in contextual dependencies. Details of this experiment are provided in the Appendix.

\section{Discussion}
We conducted a preliminary experiment applying Direct Preference Optimization (DPO)\cite{rafailov2024direct} to our models trained on SCG Instructions. Results were inconsistent, with DPO occasionally causing significant performance decreases. This may be due to the small edit distance between preferred and dispreferred responses, known to potentially harm model performance~\cite{pal2024smaug}. Performance decline was most common in datasets with shorter lengths and fewer event types, though this wasn't consistent across all models. In some cases, DPO marginally improved performance. Further investigation is needed to understand these effects. Details on this experiment are provided in the Appendix.


To additionally investigate the relationship between data complexity and model performance, we considered four dimensions: average token length, which indicates document size; the ratio of multi-word triggers, to capture the complexity of event mentions; the average number of triggers per document, reflecting event density; and the number of possible event types, representing the breadth of classification. To create a composite complexity score, we combined these factors using the L2 norm of average token length, triggers per document, multi-word trigger ratio, and event types to capture the intrinsic complexity of the data. As illustrated in Figure \ref{fig:dataset_complexity}, as dataset complexity increases, model performance generally decreases across all three models. Full details of this experiment are provided in the Appendix.

\section{Conclusion}
\label{sec:Conclusion}

In this paper, we presented a novel approach to enhance LLMs in event detection through the introduction of Semantic Causal Graphs (SCGs) and Semantic Causal Graph Instructions. We introduce SGCs as directed graphs that capture both causal relationships and contextual information within the text, providing a structured representation of events and their triggers. Our method addresses critical challenges in leveraging LLMs for event detection by utilizing these causal relationships, leading to significant improvements in performance across multiple event detection metrics.
Our comprehensive study, encompassing multiple LLMs, datasets, and training strategies, provides valuable insights into the effective use of LLMs for event detection and the retention of the LLM's reasoning capabilities. This research contributes significantly to the field of event detection using LLMs, offering a structured and effective method for improving model performance. Furthermore, the success of our approach in leveraging causal relationships suggests potential applications in other tasks where understanding causality is crucial.

\bibliography{aaai25}

\newpage

\section{Reproducibility Checklist}

\begin{itemize}
    \item This paper:
    \begin{itemize}
        \item Includes a conceptual outline and/or pseudocode description of AI methods introduced \textbf{(yes)}
        \item Clearly delineates statements that are opinions, hypothesis, and speculation from objective facts and results \textbf{(yes)}
        \item Provides well marked pedagogical references for less-familiar readers to gain background necessary to replicate the paper \textbf{(yes)}
    \end{itemize}

    \item Does this paper make theoretical contributions? \textbf{(yes)}

    \item[] If yes, please complete the list below.

    \begin{itemize}
        \item All assumptions and restrictions are stated clearly and formally. \textbf{(yes)}
        \item All novel claims are stated formally (e.g., in theorem statements). \textbf{(no)}
        \item Proofs of all novel claims are included. (yes/partial/no) \textbf{(yes)}
        \item Proof sketches or intuitions are given for complex and/or novel results. \textbf{(yes)}
        \item Appropriate citations to theoretical tools used are given. \textbf{(yes)}
        \item All theoretical claims are demonstrated empirically to hold. \textbf{(yes)}
        \item All experimental code used to eliminate or disprove claims is included. \textbf{(yes)}
    \end{itemize}

    \item Does this paper rely on one or more datasets? \textbf{(yes)}

    \item[] If yes, please complete the list below.

    \begin{itemize}
        \item A motivation is given for why the experiments are conducted on the selected datasets \textbf{(yes)}
        \item All novel datasets introduced in this paper are included in a data appendix. \textbf{(NA)}
        \item All novel datasets introduced in this paper will be made publicly available upon publication of the paper with a license that allows free usage for research purposes. \textbf{(NA)}
        \item All datasets drawn from the existing literature (potentially including authors' own previously published work) are accompanied by appropriate citations. \textbf{(yes)}
        \item All datasets drawn from the existing literature (potentially including authors' own previously published work) are publicly available. \textbf{(yes)}
        \item All datasets that are not publicly available are described in detail, with explanation why publicly available alternatives are not scientifically satisficing. \textbf{(NA)}
    \end{itemize}

    \item Does this paper include computational experiments? \textbf{(yes)}

    \item[] If yes, please complete the list below.

    \begin{itemize}
        \item Any code required for pre-processing data is included in the appendix. \textbf{(yes)}
        \item All source code required for conducting and analyzing the experiments is included in a code appendix. \textbf{(yes)}
        \item All source code required for conducting and analyzing the experiments will be made publicly available upon publication of the paper with a license that allows free usage for research purposes. \textbf{(yes)}
        \item All source code implementing new methods have comments detailing the implementation, with references to the paper where each step comes from \textbf{(partial)}
        \item If an algorithm depends on randomness, then the method used for setting seeds is described in a way sufficient to allow replication of results. \textbf{(yes)}
        \item This paper specifies the computing infrastructure used for running experiments (hardware and software), including GPU/CPU models; amount of memory; operating system; names and versions of relevant software libraries and frameworks. \textbf{(yes)}
        \item This paper formally describes evaluation metrics used and explains the motivation for choosing these metrics. (yes/partial/no)
        \item This paper states the number of algorithm runs used to compute each reported result. \textbf{(yes)}
        \item Analysis of experiments goes beyond single-dimensional summaries of performance (e.g., average; median) to include measures of variation, confidence, or other distributional information. \textbf{(no)}
        \item The significance of any improvement or decrease in performance is judged using appropriate statistical tests \textbf{(no)}
        \item This paper lists all final (hyper-)parameters used for each model/algorithm in the paper's experiments. \textbf{(yes)}
        \item This paper states the number and range of values tried per (hyper-) parameter during development of the paper, along with the criterion used for selecting the final parameter setting. \textbf{(yes)}
    \end{itemize}
\end{itemize}

\newpage

\appendix

\section{Appendix}

This appendix provides supplementary information and to support the main text of the research paper. The content of the appendix has been organized into six sections:
\begin{itemize}
\item \textbf{Appendix A: Dataset Details} - Provides information about the dataset used in the study, including details about the prompts used.

\item \textbf{Appendix B: LLMs for Event Detection} - Describes the hyperparameters, hardware, and methods used for various LLM approaches in event detection, including open-source models, APIs, full parameter fine-tuning, and RAG.

\item \textbf{Appendix C: SCG Instruction Fine-tuned LLMs vs. Deep Learning Models} - Compares SCG Instruction Fine-tuned LLMs with Deep Learning Models, detailing training methods and hyperparameters for the deep learning approaches.

\item \textbf{Appendix D: Ablation Study} - Explains the creation of the dataset used in the ablation study and reiterates the testing and inference procedures.

\item \textbf{Appendix E: DPO} - Discusses the use of the Direct Preference Optimization (DPO) method, including libraries and hyperparameters used, experimental results, and an expanded discussion of the findings.

\item \textbf{Appendix F: Data Complexity} - Presents the equation and calculations for data complexity, provides an expanded discussion on its limitations, and explains the rationale behind the selection of each factor.
\end{itemize}

\section{Appendix A: Dataset Details}
To create the SCG Instruction Datasets, we began by generating 20 variations of instructions that guide the language model to identify event triggers and classify them into corresponding event types. These variations were designed to provide the model with a diverse set of instructions, ensuring robustness and generalization in understanding and responding to event detection tasks. 
For each document or text sample from the event detection datasets, we constructed an SCG instruction using the following steps:
\begin{enumerate}
    \item Randomly select one of the 20 instruction variations to ensure diversity in the training data. All possible variations of these instructions are listed in Table \ref{tab:prompt-variations}.
    \item Append the document or text sample to the selected instruction, providing the model with the necessary context for event detection.
    \item Map the triggers and event types for the document or text sample into a structured format for all triggers and their respective event types in that document or text. 
    \item Use special tokens to demarcate the instruction and the expected output sections within the SCG data point. This helps the model distinguish between the instruction with the text of interest and the triggers and event types that it should predict during training and inference.
\end{enumerate}

\label{app:dataset}
\begin{table*}
\setlength{\tabcolsep}{3pt}
\begin{tabular}{|p{\textwidth}|}
\hline
\multicolumn{1}{|c|}{\textbf{Instruction Variations for Dataset Construction}} \\
\hline
1. As an event detection assistant, your task is to identify the event triggers in the given text. An event trigger is a word or phrase that most explicitly describes the event happening in the text. Classify each trigger into its corresponding event type from the predefined set. \\
\hline
2. In your role as an event detection assistant, find the event triggers which are words or phrases that most explicitly describe the events occurring in the text. Categorize each trigger into its respective event type. \\
\hline
3. You are an event detection assistant. Locate the event triggers, which are words or phrases that most clearly express the events in the text. Determine the corresponding event type for each trigger from the provided categories and output the trigger words along with their associated types. \\
\hline
4. Analyze the given text and pinpoint the event triggers which are specific words or phrases that most explicitly indicate the occurrence of events. As an event detection assistant, classify each event trigger into one of the predefined event types and list the triggers along with their assigned types. \\
\hline
5. In your capacity as an event detection assistant, examine the provided passage and identify the event triggers, which are key terms that most explicitly signify events taking place. For each event trigger found, specify the relevant word(s) and label it with the appropriate event type from the given set. \\
\hline
6. Read through the text and spot the event triggers which are expressions that most unambiguously represent events. As an event detection assistant, extract these event triggers and match them with their respective event types based on the predefined categories. \\
\hline
7. You are tasked with being an event detection assistant. Scan the given text to locate event triggers, which are words or phrases that most clearly denote events. For each detected event trigger, determine its event type from the provided list and output the trigger along with its corresponding type. \\
\hline
8. Go through the passage and recognize the event triggers which are terms that most explicitly indicate the presence of events. In your role as an event detection assistant, classify these event triggers into their relevant event types and generate a list containing the triggers and their assigned categories. \\
\hline
9. As an event detection assistant, identify the event triggers in the text, which are keywords that most unambiguously suggest the occurrence of specific events. Map each event trigger to one of the predefined event types and create an output featuring the triggers and their associated types. \\
\hline
10. Inspect the given text for event triggers which are words or phrases that most explicitly signal events. As an event detection assistant, categorize each discovered event trigger into its appropriate event type and produce a result that includes the trigger expressions and their corresponding types. \\
\hline
11. You are an event detection assistant. Detect the presence of event triggers which are words or phrases that most clearly describe events within the provided text. For each trigger identified, establish its event type based on the predefined categories and present the trigger along with its assigned type.\\
\hline
12. Examine the passage to uncover event triggers, which are terms that most unambiguously indicate events. In your capacity as an event detection assistant, assign each event trigger to one of the given event types and generate an output that lists the triggers and their respective categories.\\
\hline
13. As an event detection assistant, analyze the text to find event triggers which are specific words or phrases that most explicitly suggest the existence of events. Determine the event type for each trigger based on the provided categories and create a result that showcases the triggers alongside their corresponding types. \\
\hline
14. Your role is to be an event detection assistant. Study the given text and isolate the event triggers, which are expressions that most clearly imply the occurrence of events. Sort each event trigger into its designated event type and compile a list of the triggers with their assigned types. \\
\hline
15. Act as an event detection assistant and scrutinize the passage for event triggers which are indicators that most explicitly denote events. Identify the event triggers and align them with their appropriate event types based on the predefined categories. Present your findings as a list of triggers and their corresponding types.\\
\hline
16. As an event detection assistant, your objective is to pinpoint the event triggers in the text, which are words or phrases that most unambiguously signify events. Classify each event trigger into one of the given event types and generate an output that displays the triggers alongside their associated types. \\
\hline
17. In your function as an event detection assistant, review the provided text and highlight the event triggers which are terms that most clearly denote events. Assign each event trigger to its relevant event type and produce a result that showcases the triggers and their corresponding categories. \\
\hline
18. You are an event detection assistant tasked with identifying event triggers which are words or phrases that most explicitly describe events within the given text. Determine the event type for each trigger based on the predefined set and create an output that lists the triggers along with their assigned types. \\
\hline
19. As an event detection assistant, evaluate the passage to discover event triggers, which are expressions that most unambiguously indicate events. Categorize each event trigger into one of the provided event types and compile a list that includes the triggers and their respective types. \\
\hline
20. In your role as an event detection assistant, examine the text to locate event triggers which are specific words or phrases that most explicitly imply events. Map each event trigger to its appropriate event type based on the given categories and generate a result that presents the triggers and their corresponding types. \\
\hline
\end{tabular}
\caption{All instruction variations used in dataset creation}
\label{tab:prompt-variations}
\end{table*}

\section{Appendix B: LLMs for Event Detection}
\label{app:llms}
\textbf{LLMs and Hardware:} For open-source LLMs, we used a LoRA setup with a rank of 256 and an alpha of 256 during training. Our training process used 6 epochs with a batch size of 16. We used the AdamW optimizer in 8-bit precision, with a learning rate of 5e-5 and a cosine learning rate scheduler with a warmup ratio of 0.1. In the full parameter fine-tuning experiment, we used a batch size of 1 and trained for 6 epochs. For inference, we used the default sampling parameters from HuggingFace, with both temperature and top-p set to 1.0. The hardware used for running open-source LLMs was an NVIDIA A100 80GB GPU.
Similarly, for API-based proprietary LLMs, we used the default sampling parameters provided by OpenAI, with a temperature of 1.0 and top-p of 1.0.

\textbf{Retrieval-Augmented Generation Inference:} For our RAG implementation, we chose the 'all-MiniLM-L6-v2' sentence-transformers model ~\cite{reimers-2019-sentence-bert} from HuggingFace as our embedding model which maps sentences and paragraphs to a 384-dimensional dense vector space. This selection was based on its efficient performance in sentence and paragraph embedding tasks. 

\section{Appendix C: SCG Instruction Fine-tuned LLMs vs. Deep Learning Models}
\label{app:scg}
For the deep learning methods, the following are the models evaluated and their training hyperparameters:

TagPrime-C~\cite{hsu2023tagprime}: A sequence tagging model that appends priming words about the information of the given condition (such as an event trigger) to the input text. Training hyperparameters: 90 epochs, batch size of 6, and learning rate of 1e-5.

CEDAR~\cite{li2023glen}: A multi-stage cascaded event detection model designed to address the challenges of large ontology size and distant-supervised data. Training hyperparameters: 5 epochs, batch size of 128, and learning rate of 1e-5.

DEGREE~\cite{hsu2022degree}: A data-efficient model that learns to summarize the events mentioned in a text into a natural sentence that follows a predefined pattern. Training hyperparameters: 45 epochs, batch size of 32, and learning rate of 1e-5.

\section{Appendix D: Ablation Study}
\label{app:ablation}
For our ablation study, we created a dataset by modifying the text fields of original event detection test sets using Claude 3.5 Sonnet~\cite{anthropic_claude35sonnet_2024}. The aim was to alter the context while preserving the event triggers, allowing us to test model robustness against contextual changes. Claude 3.5 Sonnet was provided with a system prompt (detailed in Table \ref{tab:system-prompt}) instructing it to change aspects such as entities, locations, dates, times, and other contextual details, while maintaining the original event triggers. To ensure trigger preservation, we supplied the golden label triggers with each prompt and implemented a verification process. If Claude failed to include all original triggers, we repeated the generation process until successful. This methodology allowed us to create variations that challenged the models' ability to detect events across different contexts while keeping the core event information intact.

\begin{table*}
\begin{tabular}{p{\textwidth}}
\toprule

You are an AI assistant tasked with modifying text for event detection datasets to make variations of the original text. Your job is to change only entities, locations, dates, times, and similar specific details in the given text. Do not alter the overall structure, events, or meaning of the text. Maintain the same writing style and tone. Your output should be the modified text only, without any explanations or additional comments. \\

Rules: \\
1. Change names of people, organizations, and locations. \\
2. Modify dates and times, but keep them realistic and consistent with the events described. \\
3. Alter specific numbers (e.g., ages, quantities) slightly, but keep them plausible. \\
4. Do not change the events, their types, or their triggers. \\
5. Maintain the same paragraph structure and quotations (if any). \\
6. Ensure the modified text remains coherent and logical. \\
7. Keep all original trigger phrases intact and in the same context. \\
8. Other than the modifications, keep all other input text exactly the same. \\
9. If a sentence or phrase does not contain any entities, locations, dates, or times to be changed, leave it completely unchanged. \\
10. Ensure that all user-provided trigger words/phrases from the original text are included in the modified text in their original context. \\

Remember, the goal is to create a subtle variation of the original text while preserving its core structure and meaning. Be extremely careful not to alter anything beyond the specific elements mentioned in the rules. \\
\bottomrule
\end{tabular}
\caption{System prompt used for context modification of test sets}
\label{tab:system-prompt}
\end{table*}

\section{Appendix E: DPO}
\label{app:dpo}
We implement Direct Preference Optimization~\cite{rafailov2024direct} using the HuggingFace TRL library~\cite{vonwerra2022trl}, leveraging LoRA for efficient parameter tuning. For our LoRA setup, we configure a rank of 64 and an alpha of 64. to create preference pairs, we use the SCG-instruct model's incorrect responses on the development set of each event detection dataset as dispreferred responses and the corresponding correctly labeled responses as the preferred responses. Since the model has already been fine-tuned to the task, we only train for one epoch to allow DPO to make slight adjustments and bring the model closer to optimal performance. We conduct training with a batch size of 2, utilizing the AdamW optimizer in 8-bit precision. The learning rate is set to 5e-6, and we employ a cosine learning rate scheduler. For DPO-specific parameters, we set $\beta$ to 0.1.

We found that DPO improved or made little change to the performance of the instruction-tuned models when the event detection dataset being trained on was more complex, meaning that those datasets had the structure of having text sequences that were multiple sentences and multiple triggers/event types. In contrast, for datasets that did not have data samples of multiple sentences and multiple events per text sequence, applying DPO on top caused the performance to drop significantly. The full results are shown in Table \ref{tab:DPO_Comparison}

\begin{table*}[]
\centering
\resizebox{.99\textwidth}{!}{
\begin{tabular}{@{}l|ccc|ccc|ccc|ccc|ccc|ccc@{}}
\toprule
\multirow{4}{*}{Model}        & \multicolumn{3}{c|}{\multirow{3}{*}{MLEE}} & \multicolumn{3}{c|}{\multirow{3}{*}{FewEvent}} & \multicolumn{3}{c|}{\multirow{3}{*}{M$^2$E$^2$}} & \multicolumn{3}{c|}{\multirow{3}{*}{CASIE}} & \multicolumn{3}{c|}{\multirow{3}{*}{MAVEN}} & \multicolumn{3}{c}{\multirow{3}{*}{Average}} \\
                              & \multicolumn{3}{l|}{}                      & \multicolumn{3}{l|}{}                           & \multicolumn{3}{l|}{}                      & \multicolumn{3}{l|}{}                       & \multicolumn{3}{l|}{}                       & \multicolumn{3}{l}{}                         \\
                              & EC           & TI           & TC           & EC             & TI             & TC            & EC           & TI           & TC           & EC            & TI           & TC           & EC            & TI           & TC           & EC            & TI            & TC           \\ 
\midrule
Mistral-LoRA-SCG Instruct + DPO
& 87.14  & 49.08  & 42.78 
& 14.77  & 1.46   & 1.46  
& 92.62  & 79.73  & 79.30 
& 83.67  & 43.01  & 42.30 
& 4.05   & 5.20   & 0.23  
& 56.45  & 35.70  & 33.21
\\
Llama 3-LoRA-SCG Instruct + DPO
& 90.36  & 61.99  & 55.76
& 6.36   & 41.51  & 2.78  
& 77.49  & 67.44  & 66.86 
& 96.71  & 50.36  & 49.94 
& 42.89  & 36.55  & 19.56 
& 62.76  & 51.57  & 38.98
\\
Gemma-LoRA-SCG Instruct + DPO
& 84.54  & 51.93  & 44.00 
& 63.40  & 27.28  & 26.23
& 11.21  & 10.40  & 10.39 
& 87.09  & 36.39  & 34.86 
& 52.54  & 32.10  & 17.55 
& 59.76  & 31.62  & 26.61
\\
\bottomrule
\end{tabular}}

\caption{Performance comparison of various models with DPO on multiple datasets. F1 scores are reported.}
\label{tab:DPO_Comparison}
\end{table*}

\section{Appendix F: Data Complexity}
\label{app:complexity}
While investigating the performance of our models, we recognized the importance of quantifying dataset complexity to better understand the relationship between data characteristics and model performance. Thus, we created a composite complexity score based on four key dimensions that we believed captured essential aspects of dataset difficulty in the context of event detection tasks. The four dimensions we chose for our complexity analysis were: 1) average token length, 2) ratio of multi-word triggers, 3) average number of triggers per document, and 4) number of possible event types. Each of these dimensions was selected to represent a different facet of complexity in event detection tasks.

Average token length serves as a proxy for document size, which can affect the model's ability to maintain context over longer sequences. The ratio of multi-word triggers captures the complexity of event mentions themselves, as multi-word triggers often require more sophisticated understanding than single-word triggers. The average number of triggers per document reflects event density, which can impact the model's ability to distinguish between multiple events in close proximity. Finally, the number of possible event types represents the breadth of classification, indicating the diversity of events the model must learn to identify.

To create a composite complexity score that incorporates all these factors, we utilized the L2 norm of these four dimensions. The complexity score C is computed as follows (where ATL is the average token length, MTR is the multi-word trigger ratio, TPD is the average number of triggers per document, and ET is the number of event types):

\begin{equation}
C = \sqrt{(ATL)^2 + (MTR)^2 + (TPD)^2 + (ET)^2}
\end{equation}

Each metric and the computed complexity score are shown in Table \ref{tab:dataset_complexity-metrics}.


\begin{table}[h]
\centering
\resizebox{.99\columnwidth}{!}{
\begin{tabular}{@{}l|r|r|r|r|r@{}}
\toprule
\textbf{Dataset} & \textbf{ATL} & \textbf{TPD} & \textbf{ET} & \textbf{MTR} & \textbf{C} \\
\midrule
M2E2 & 30.33 & 1.03 & 8 & 0.04 & 31.38 \\
FewEvent & 35.55 & 1.00 & 100 & 0.14 & 106.14 \\
MAVEN & 29.94 & 2.48 & 168 & 0.04 & 170.67 \\
MLEE & 301.72 & 23.65 & 29 & 0.07 & 304.03 \\
CASIE & 316.62 & 5.81 & 5 & 0.69 & 316.71 \\
\bottomrule
\end{tabular}}
\caption{Dataset Complexity Metrics and Scores.}
\label{tab:dataset_complexity-metrics}
\end{table}

While these four dimensions provide a solid foundation for assessing dataset complexity in event detection tasks, it's important to acknowledge the limitations of this approach. The chosen dimensions may not fully capture the complexity of datasets from all domains or account for other factors that are difficult to quantify. For instance, this metric doesn't directly account for linguistic complexity, contextual dependencies, or the subtlety of event descriptions that might require world knowledge or complex reasoning to detect. The relative importance of each dimension might also vary depending on the specific task or domain, which our equal-weighted approach doesn't account for.
Despite these limitations, our complexity score provides a useful starting point for comparing datasets and understanding how different aspects of data complexity might influence model performance. Future work could explore additional dimensions or alternative weighting schemes to create more comprehensive or domain-specific complexity metrics.

\end{document}